\documentclass[conference,compsoc]{IEEEtran}
\usepackage{hyperref}

\ifCLASSOPTIONcompsoc
  \usepackage[nocompress]{cite}
\else
  \usepackage{cite}
\fi
\ifCLASSINFOpdf
  \usepackage[pdftex]{graphicx}
\else
\fi
\begin{document}

%

\title{Grandma Karl is 27 years old -- \\ research agenda for pseudonymization of research data}





\author{\IEEEauthorblockN{Elena Volodina and Simon Dobnik}
\IEEEauthorblockA{University of Gothenburg, Sweden\\
name.surname@gu.se}
\and
\IEEEauthorblockN{Therese Lindström Tiedemann}
\IEEEauthorblockA{University of Helsinki, Finland\\
therese.lindstromtiedemann@helsinki.fi}
\and
\IEEEauthorblockN{Xuan-Son Vu}
\IEEEauthorblockA{Umeå University, Sweden\\
sonvx@cs.umu.se}}


%


\maketitle

\begin{abstract}
Accessibility of research data is critical for advances in many research fields, but textual data often cannot be shared due to the  personal and sensitive information which it contains, e.g names or political opinions. General Data Protection Regulation (GDPR) suggests pseudonymization as a solution to secure open access to research data, but we need to learn more about pseudonymization as an approach before adopting it for manipulation of research data. This paper outlines a research agenda within pseudonymization, namely need of studies into the effects of pseudonymization on unstructured data in relation to e.g. readability and language assessment, as well as the effectiveness of pseudonymization as a way of protecting writer identity, while also exploring different ways of developing context-sensitive algorithms for detection, labelling and replacement of personal information in unstructured data. 
The recently granted project on pseudonymization `Grandma Karl is 27 years old'\footnote{\url{https://spraakbanken.gu.se/en/projects/mormor-karl}} addresses exactly those challenges.
\end{abstract}


%
\IEEEpeerreviewmaketitle

\section{Introduction}
Access to quality language data is a prerequisite for research in digital humanities and applied linguistics as well as for the development of Artificial Intelligence (AI) and Natural
Language Processing (NLP) based approaches. 
However, accessible data becomes a challenging target when personal data is involved. This is very true of data used in digital humanities and applied linguistics, for example, language learner data often contains tasks phrased so that the tasks – directly or indirectly – elicit explicit personal information, e.g. ``Introduce yourself'' or ``Describe the best or worst day of your life'' (cf. Stemle et al. \cite{21@stemle2019working}).

\begin{figure*}[t!]
    \centering
    \includegraphics{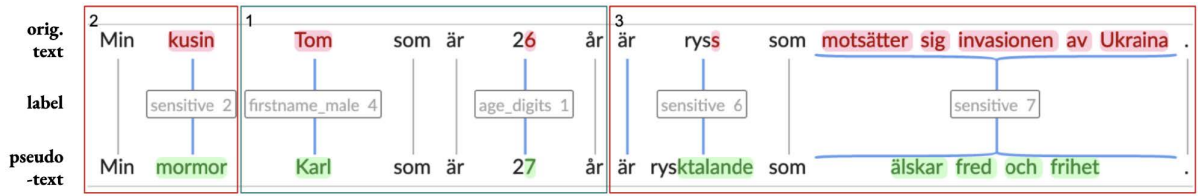}
    \caption{Current state of automatic pseudonymization of freely written texts in Swedish [27]. (1) is a solved issue by detection using regular expressions or Named Entity Recognition at the token level. (2) is a partly solved task where
token-level or minimal context is sufficient for detection, but insufficient for pseudonymization of personal identifiers. World
knowledge and semantic constraints (e.g., gender, age) need to be encoded. (3) is a (political/religious/…) stance detection,
and is a Natural Language Understanding task, requiring new techniques both for detection and to find appropriate
paraphrases as pseudonyms.
Gloss: Top level -- simulated original text: My cousin Tom who is 26 years old is a Russian who is against the assault on
Ukraine. Middle level -- pseudo-labels. Bottom level -- pseudonymized text: automatically selected pseudonyms for \cite{1@kwaik2020arabic} and
\cite{2@adelani2020privacy}; and manually added pseudonyms for \cite{3@adouane2017identification}: My *grandma Karl who is 27 years old, is Russian-speaking and loves peace
and freedom.}
    \label{fig:pseudo_exe}
\end{figure*}

The General Data Protection Regulation (GDPR) \cite{gdpr2018} lists pseudonymization as a method to conceal a
real person behind the data -- as well as a way to show GDPR compliance with requirements such
as “data protection by design”. However, as a method pseudonymization still raises several open questions:
\begin{itemize}
\item what personal information must be manipulated to achieve an acceptable level of protection
without over-manipulating research data? 
This means that we need to 
evaluate effectiveness of pseudonymization as a method to protect writers' privacy, 
i.e. a proof that pseudonymization really works;
\item how to automatically detect and pseudonymize Personally Identifiable Information (PII), both
direct identifiers and indirect identifiers (so called quasi-identifiers), especially the ones containing
spelling errors (\texttt{*stockhulm} `Stockholm') or other linguistic, semantic and contextual challenges
(e.g. see Fig.\ref{fig:pseudo_exe})? 
This means that algorithms must be sensitive to linguistic constraints and should not
corrupt research data more than 
necessary 
to protect personal information;
\item how does pseudonymization 
affect
data, and consequently, theoretical
conclusions and practical applications based on such data? 
That is, pseudonymized data should not only provide data where personal identifiers have been efficiently removed, it should be functional and usable 
for research (e.g. in applied linguistics or social sciences) despite
being manipulated.
\end{itemize}

\begin{figure*}[t!]
    \centering
    \includegraphics{pseudo\_types.jpg}
    \caption{Mocked original learner text and several ways of masking potentially sensitive information.}
    \label{fig:pseudo_types}
\end{figure*}

We argue that the current state of (knowledge about) pseudonymization is insufficient and requires thorough research. Lison et al. \cite{Lison:2021ab} describes the current state of the art in this area 
and identifies several challenges related to the current methods used. In our view, further work is required in relation to exploring the effects of these methods in different contexts of data and differences required in processing languages other than English. 

Furthermore, language data which is explicitly collected for research  on language, e.g. for research on language development or language assessment, complicates these issues even further since each linguistic token is part of the actual data under study. Nevertheless, there is still a need to 
pseudonymize such data as well, not only for research purposes but also in relation to assessment itself where there have been attempts over the last decades to make assessment anonymous, 
an issue which is particular complex when the person being assessed can be identified by contextual clues 
and not only by a name or identifier in the meta information attached to the test. 

With this paper we introduce the research agenda pursued in the project \textit{Grandma Karl is 27 years old: Automatic pseudonymization of research data}, which focuses on Swedish research data using learner corpora as the main focus for the study.



\section{Previous work and open questions}

\subsection{NLP and pseudonymization}
\label{nlp}

Text de-identification and pseudonymization are two ways of protecting the privacy of individuals by
transforming their personal data. 
\textit{De-identification} is the process of removing
all information that could identify an individual (see Alt.1-2 in Fig.\ref{fig:pseudo_types}). Text de-identification, (a.k.a sanitization), is a well established task and is
required for processing data in highly sensitive domains (e.g. for patient health records). 
\textit{Pseudonymization} is the process of replacing an
individual’s personal data with a pseudonym, which is 
not related 
to the original data
 (see Alt.3-4 in Fig.\ref{fig:pseudo_types}). 
In healthcare,
pseudonymization allows the data to be associated with a patient only under specified and controlled
circumstances \cite{16@neubauer2011methodology}. 





Generally, there are two types of de-identification: deterministic and probabilistic. Deterministic
de-identification is a process in which each instance of a potentially identifying element is replaced
with the same 
non-related identifier (e.g. ****). 
Probabilistic de-identification is a process in which each instance of a
potentially identifying element is replaced with a different 
non-related identifier, e.g. OBJ63843.

A large number of de-identification methods have been proposed 
(\cite{5@dernoncourt2017identification,15@meystre2010automatic}). While these methods are
applied in a number of domains with a focus on privacy protection, we see a need for a different approach, namely an
optimization for a good trade-off between de-identification and preservation of the semantic meaning of the texts 
and as much of the linguistic characteristics of the text as possible, as ways of facilitating related research in linguistics, digital humanities, and in other fields. Therefore, we see a need to maximize the
de-identification while minimizing e.g. the change in the semantic and pragmatic meaning of the original texts, the readability of the text, or the level of proficiency one would assume that the writer has. This objective is extremely challenging since many of the de-identification methods are designed to
remove/replace specific identifiers. Adelani et al. \cite{2@adelani2020privacy} reported that using a state-of-the-art deep learning model to
train natural language understanding tasks by fine-tuning BERT on de-identified texts imposes a considerable 
destructive effect on 
its
performance. 
Thus, new approaches using so-called semantic preserving de-identification (a.k.a SP-de-identification) are in demand. 

Since the introduction of the GDPR legislation \cite{gdpr2018} we have seen an increased interest in the NLP
community in how to deal with automatic anonymization (another term for `de-identification' adopted in GDPR)
and pseudonymization (\cite{9@kondratyev2020data,19@rocher2019estimating,30@xia2021enabling}). 
However, most of the
literature on the topic deals with medical data \cite{12@marimon2019automatic} and the English language remains the main focus \cite{17@pilan2022text}. 

Notably,
there are to our knowledge no known studies about how to identify unstructured personal/sensitive information in a
narrative (e.g. \texttt{When I was five I fell off a mango tree, and I limp since then}),
whether it is important to project errors to pseudonyms (e.g. \texttt{*stockhulm → *gothenberg};
`Stockholm' → `Gothenburg') (cf. Stemle et al. \cite{stemle2019clarin})
and how to estimate the semantic and contextual appropriateness
of a pseudonym in the current context without introducing any new errors or skewing the interpretation of the text (see Fig.\ref{fig:pseudo_exe}).

\subsection{Effects of pseudonymization on research data}
\label{effects}

Since most previous research on pseudonymization has focused on the medical domain, we are more  or less in the "unknown waters" when it comes to pseudonymization in other domains. Thus,
if we consider unstructured research data such as learner essays written 
for purpose of training writing skills or for assessing a learner's language proficiency level, what types of pseudonymization might be necessary and how might this affect the data in relation to the objectives for which these essays are usually written?
To minimise the risk that PIIs
might reveal authors' identity,
all
identifiers in 
essays 
need to be detected, masked and 
eventually replaced. 
Thus,
pseudonymization includes identification of personal information that can relate to the 
author 
(e.g.
My name is \texttt{Ali}) and classification of that information to a certain predefined type (e.g. My name is
\texttt{first name}). Each information type can then be replaced in a systematic way to reproduce a
``natural'' text to retain the reading flow (e.g. My name is \texttt{Robert}, where the original first name is
replaced randomly by another first name). Megyesi et al. 
\cite{megyesi2018anonym} list several ways to mask the sensitive
information in the pseudonymization process, among others by using a placeholder (\texttt{Poland →
@A-country}, Alt.2 in Fig.\ref{fig:pseudo_types}); through substitution (\texttt{Poland → Greece}, Alt.3 and 4 in Fig.\ref{fig:pseudo_types}); by making
text noisy (\texttt{Poland → Europe}); or by completely removing a text segment (\texttt{Poland → ****}, Alt.1 in Fig.\ref{fig:pseudo_types}).


Since automatic approaches to pseudonymization are only starting to arise, there are no known studies
on the effect of different types of pseudonymization on readability (cf. Fig.\ref{fig:pseudo_types}, Alt.1--4), on the utility
of pseudonymized data for different research questions, such as language assessment, on the general
accuracy of theoretical generalisations (e.g. about the linguistic competence of the learners, cf. Volodina et al. \cite{26@volodina2022reliability}) or
on the general attitudes that pseudonyms may provoke (cf. Fig.\ref{fig:pseudo_types}, Alt.3 and 4 for cultural contexts). However, a small study by Aldrin \cite{aldrin2017assessing} 
has shown that changing the names which appear in essays written for first language assessment can interfere with the assessment of language proficiency when names are picked from different ethnic or social backgrounds.
On a
more theoretical side, the most active opponent to any kind of text annotation, Sinclair \cite{20@sinclair2004trust}, claimed
that \textit{any} data annotation destroys the original data, since meddling with the original data “imposes
theoretical preconceptions on the data [\ldots] and destroys the integrity of the text” which echoes the first
annotation maxim by Leech \cite{11@leech1993corpus} that claims that the "pure" (original) corpus should be recoverable
after annotation. This sets further needs to explore the effects of pseudonymization on the original data,
especially if pseudonymization is adopted as a standard way of pre-processing research data, so that
we can find an acceptable trade off between data privacy protection and data utility preservation.

\subsection{Assessment of the re-identification risks}
\label{re-id}

The former concept of anonymization was binary: either data was seen as anonymous or not \cite{22@sweeney2000simple}.
Removing explicit identifiers, for example, name, date of birth and telephone number, was considered
sufficient. However, identifiability is relative and contextual, and seemingly impersonal data points --
that on their own apply to many people -- can, if taken together, identify people uniquely. Sweeney
\cite{22@sweeney2000simple} has shown that by providing date of birth, zip code and gender, 87\% of the US population could
be uniquely identified. When these identifiers appear in unstructured texts, e.g. in personal stories,
blogs or language learner essays, the task of detecting and masking them becomes even more urgent
and challenging.

On completion of pseudonymization or other data-protective anonymization techniques on a dataset
for release, techniques known as motivated intruder tests or re-identification tests can be used (e.g. \cite{19@rocher2019estimating,24@tudor2014intruder}) to test the effectiveness of the privacy protection used.
This helps to assess the risk of a person being identified in a pseudonymized dataset using all sources
of information at hand. The risk assessment based on that can, then, regulate how many identifiers that
actually need to be pseudonymized, and the level of access that can be granted to the public or
researchers. Unfortunately, very little has been published about this type of testing. Tudor et al. \cite{24@tudor2014intruder}
suggest that there is surprisingly little known about how trivial it is to identify individuals behind
various points of information, what techniques “intruders” might resort to in order to disclose private
people, and how to realistically set protection (as opposed to over-protection) levels of data access.

We acknowledge the need to combine automatic (cf. Rocher et al. \cite{19@rocher2019estimating}) and potentially manual (cf. Tudor et al. \cite{24@tudor2014intruder}) approaches to
re-identification to identify absolutely necessary entities for pseudonymization, this way addressing both purely academic and practical needs. Most importantly, we argue that what is generally
lacking in discussions of pseudonymization is how to find a compromise and unite two equally
important aims: (1) the aim to protect the person behind the data and (2) the aim to base research on
original (e.g. non-manipulated or minimally manipulated) data (cf. Sinclair \cite{20@sinclair2004trust}).

\section{Tying it all together}
\label{tying}

To tie it all together, we are working on pseudonymization of learner corpora within the project \href{https://spraakbanken.gu.se/en/projects/mormor-karl}{GRANDMA KARL is 27 YEARS OLD}, experimenting with the three questions outlined above: context-aware algorithms, effects of pseudonymization and effectiveness of pseudonymization. The current work is in its initial stage and takes its start in Volodina et al. \cite{28@volodina2019swell} which focused on pseudonymization of
learner essays, introducing a set of categories, principles and procedures for manual pseudonymization \cite{14@megyesi2018learner} for which a special tool, SVALA, 
was developed \cite{29@wiren2019svala} using rule-based automatic pseudonymization.
Results of that pilot experiment showed that the pseudonymizer service captures certain personal
information in narratives, argumentative and instructional texts, but fails in investigative and
evaluative genres. Deviating use of language has proven to be a major obstacle to correctly detect
personal identifiers, e.g. lacking capitalization, misspellings, homonymy between proper names,
common nouns and pronouns, etc. The pseudonymization 
step, however, was significantly more
difficult, since numerous common knowledge and semantic constraints could not be met with the
rule-based approach, which instead corrupted the data, often rendering it nonsensical (e.g. Fig.\ref{fig:pseudo_exe} \cite{2@adelani2020privacy}).




The pilot experiment has revealed a number of challenges, both generally linguistic ones and specific
to second language writing. There are also clear methodological challenges that have become obvious,
such as risks of introducing errors through pseudonyms (Fig.\ref{fig:pseudo_exe}), the readability of the resulting text
(Fig.\ref{fig:pseudo_types}), and others.

We are investigating the challenges outlined above using Swedish learner written texts \cite{28@volodina2019swell} given that
this data versatility has high potential to generalize to other personal-oriented domains, e.g. social
networks, court cases or similar. 
Several characteristics make language learner texts appropriate for studying the outlined questions:
\begin{itemize}
    \item learner texts are often personal in nature and have varied topics;
    \item they have associated socio-demographic information that increases the risks of re-identification;
    \item they abound in errors and non-standard language making automatic pseudonymization a challenge.
\end{itemize}

Besides, learner essays contain texts whose language quality is critical for real-life applications (e.g.
error correction) or research analysis (e.g. language acquisition). Hence, it is important that we
“handle it with care”, i.e. do not corrupt the data unnecessarily for it to continue being useful for,
among others, Natural Language Processing (NLP) and second language acquisition (SLA) research. Learner
essays may contain sensitive information since learners frequently come from backgrounds where they
have left the country for political reasons and therefore it is crucial that their identity is not revealed,
and political, religious or other stances are masked. Most of the realities of a learner text apply to
social media texts or other personally-oriented types of texts (e.g. court records), which makes learner
essays an excellent material to develop pseudonymization for.

Besides, the solutions for the outlined problems 
must generalize to other research domains which is why the taxonomy in Megyesi et al. \cite{14@megyesi2018learner} is being extended to cover categories specific for other domains to cover diseases, nicknames and personal handles, diets and similar, bearing in mind that identification of 
data subjects 
does not always have to be followed by replacement. Research into which personal identification that needs replacement might also be partly context-dependent, genre-depended, and in relation to research, subject-related.  
Luckily, the need for this type of work has become central to many funding agencies, and a number of projects have arisen, such as \href{https://cleanup.nr.no}{The CLEANUP Project} \cite{papadopoulou-etal-2022-neural} and \href{https://spraakbanken.gu.se/en/projects/mormor-karl}{GRANDMA KARL is 27 YEARS OLD}
and we are likely to see more publications on this topic in the near future.

In our future work we will approach the task from different perspectives: (i) data collection and annotation, (ii) linguistic analysis of the domain and its relation to other related domains, pseudonymization and effects on semantics, (iii) language modelling, computational representation of meaning and evaluation of inference, (iv) data-privacy and privacy-preserving methods. Different workpackages will inform each other and hence research will be iteratively refined based on the findings of evaluation from each cycle. Our preliminary work and experience has identified that pseudonymization is a complex phenomenon and all these aspects must be considered together to ensure sufficient quality of the resulting research data.

\ifCLASSOPTIONcompsoc
  \section*{Acknowledgments}
\else
  \section*{Acknowledgment}
\fi

We thank the Swedish Research Council for its funding nr 2022-02311 for the project \textit{Grandma Karl is 27 years old: Automatic pseudonymization of research data} for years 2023-2029.



%



\bibliographystyle{IEEEtran}
\bibliography{mormor}

\end{document}